\documentclass[letterpaper]{article} 
\usepackage{aaai2026}  
\usepackage{times}  
\usepackage{helvet}  
\usepackage{courier}  
\usepackage[hyphens]{url}  
\usepackage{graphicx} 
\usepackage{natbib}  
\usepackage{caption} 
\usepackage{multicol}
\usepackage{multirow}
\usepackage{rotating}
\usepackage{subfigure}
\usepackage{booktabs}
\usepackage{amsmath}
\usepackage{amssymb}
\usepackage{xcolor}
\usepackage{algorithm}
\usepackage{algorithmic}

\usepackage{newfloat}
\usepackage{listings}
\DeclareCaptionStyle{ruled}{labelfont=normalfont,labelsep=colon,strut=off} 
\floatstyle{ruled}
\newfloat{listing}{tb}{lst}{}
\floatname{listing}{Listing}
\title{HINTS: Extraction of Human Insights from Time-Series \\ Without External Sources}
\author{
    Sheo Yon Jhin,
    Noseong Park
}
\affiliations{
    KAIST\\
    sheoyon.jhin@kaist.ac.kr, noseong@kaist.ac.kr

    Daejeon, South Korea \\
   
}

\begin{document}

\maketitle

\begin{abstract}
Human decision-making, emotions, and collective psychology are complex factors that shape the temporal dynamics observed in financial and economic systems. Many recent time series forecasting models leverage external sources (e.g., news and social media) to capture human factors, but these approaches incur high data dependency costs in terms of financial, computational, and practical implications. In this study, we propose HINTS, a self-supervised learning framework that extracts these latent factors endogenously from time series residuals without external data. HINTS leverages the Friedkin-Johnsen (FJ) opinion dynamics model as a structural inductive bias to model evolving social influence, memory, and bias patterns. The extracted human factors are integrated into a state-of-the-art backbone model as an attention map. Experimental results using nine real-world and benchmark datasets demonstrate that HINTS consistently improves forecasting accuracy. Furthermore, multiple case studies and ablation studies validate the interpretability of HINTS, demonstrating strong semantic alignment between the extracted factors and real-world events, demonstrating the practical utility of HINTS.
\end{abstract}


\section{Introduction}

\begin{figure}[h]
\centering
{{\includegraphics[width=1.0\columnwidth]{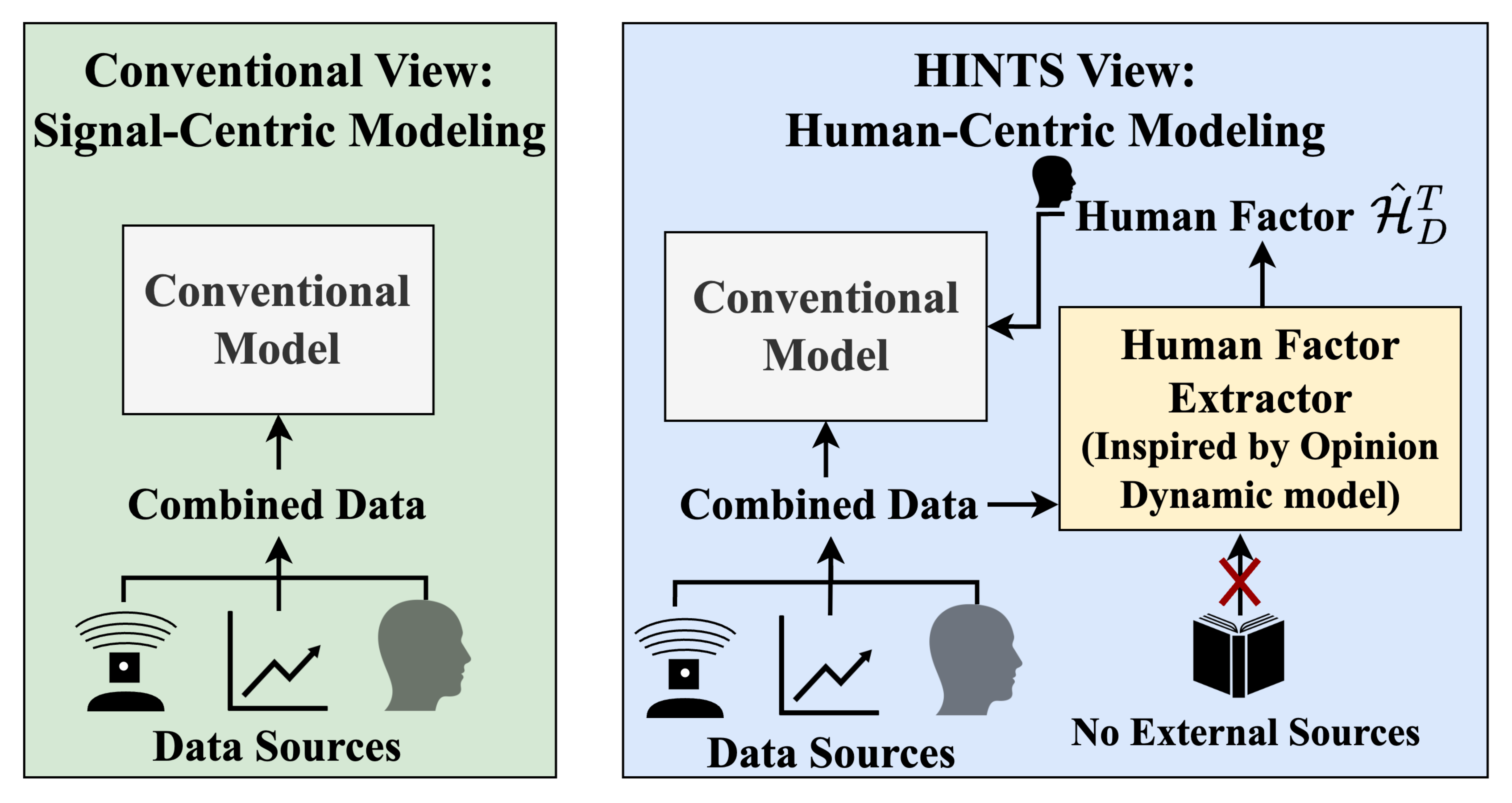}}} 
\caption{Comparison between conventional signal-centric modeling and our proposed human-centric approach. While conventional models learn directly from combined data, HINTS explicitly extracts human-driven latent factors from raw combined data alone --- without relying on any external sources such as news. Our Human Factor Extractor, inspired by opinion dynamics, identifies hidden human influence embedded in the signal and incorporates it into the conventional forecasting model, offering a structured and interpretable path from data to decision.}
\label{fig:teaser}
\end{figure}

Time-series data are pervasive across critical societal domains such as economics, finance, and transportation, where accurate forecasting is essential for understanding and responding to complex real-world dynamics. These sequences originate from diverse sources, including physical sensors, administrative records, and human-driven behaviors. For instance, financial time series contain structured variables like trading volume or corporate fundamentals, yet their outcomes reflect latent socio-economic forces such as market sentiment and investor psychology. Similarly, traffic data ultimately reflect human activity. From the moment of data generation, these human-driven influences are embedded in time-series signals like fingerprints.

Given this inherent characteristic of time series, many recent studies have sought to improve forecasting performance by incorporating external sources like news and social media~\cite{okawa2022predicting,wang2024news}. However, this approach incurs significant financial and temporal costs for collecting and refining external data, while also consuming additional computational resources for their learning and application.

We reconsider this research direction and propose a new approach that focuses on endogenous signals within the time series, rather than relying on external data. Our core hypothesis is that the influence of external factors (e.g., news, public opinion) utilized by existing studies is already reflected in the original time-series data. Specifically, we focus on the residual component, which conventional decomposition models often dismiss as random noise. Prior studies~\cite{shiller1981stock,de1990noise,neuman2023unveiling} support our hypothesis by demonstrating that so-called "unpredictable" residuals are not merely random but can encode structured behavioral dynamics linked to investor sentiment or policy shocks.

Grounded in this background, we propose HINTS, a two-stage self-supervised learning framework that endogenously extracts latent patterns, termed the 'Human Factor', from time-series residuals and connects them to prediction.
HINTS consists of (i) a Human Factor extraction stage and (ii) a prediction stage based on the extracted Human Factor. In the first stage, we leverage the Friedkin-Johnsen (FJ) opinion dynamics model as a structural inductive bias to extract the Human Factor from the residuals. This process learns to represent latent behavioral patterns by capturing the core elements of the FJ model: social influence, self-memory, and individual bias. In the second stage, the learned Human Factor is integrated into existing forecasting models as an attention mechanism. This enhances prediction accuracy while simultaneously providing interpretable insights into the underlying human behaviors.

Our key contributions are as follows:
\begin{enumerate}
\item We reconceptualize time-series residuals as carriers of human-driven dynamics and propose a self-supervised learning framework to extract these latent patterns without external data.
\item We introduce a novel methodology for extracting the Human Factor grounded in the Friedkin-Johnsen opinion dynamics model.
\item We demonstrate that integrating our proposed Human Factor into three state-of-the-art (SOTA) backbone models consistently improves forecasting accuracy and interpretability across nine real-world datasets.
\end{enumerate}



\section{Related Work}
\subsection{Opinion Dynamics and Sociological Modeling}

Opinion dynamics models have long been studied as mathematical frameworks for describing how individual beliefs evolve under social influence. Two of the most influential models in this domain are the DeGroot model and the Friedkin–Johnsen (FJ) model.

\paragraph{DeGroot Model}
The DeGroot model~\cite{degroot1974reaching} represents a consensus-seeking process in which each agent updates their opinion based on a weighted average of their neighbors' opinions. Formally, the update rule is given as:
\begin{align}
    x_i(t+1) = \sum_{j=1}^N w_{ij} x_j(t),
\end{align}
where $x_i(t)$ is the opinion of agent $i$ at time $t$, and $w_{ij}$ is the influence weight from agent $j$ to agent $i$. The matrix $W = [w_{ij}]$ is row-stochastic (i.e., each row sums to 1), ensuring convergence under mild conditions. While simple, this model assumes agents fully assimilate social information over time and always reach consensus, which may not hold in real-world social systems.

\paragraph{Friedkin–Johnsen Model}
The Friedkin–Johnsen (FJ) model~\cite{friedkin1990social} generalizes DeGroot by introducing the notion of intrinsic belief or stubbornness. Each agent maintains a latent intrinsic opinion $s_i$ and expresses a public opinion $z_i(t)$ that evolves under social pressure. The update rule incorporates both social averaging and self-retention:
\begin{align}
    z_i(t+1) = \lambda_i \sum_{j=1}^N w_{ij} z_j(t) + (1 - \lambda_i) s_i,
\end{align}
where $\lambda_i \in [0,1]$ is the susceptibility of agent $i$ to social influence. This model allows for persistent disagreement and opinion polarization, better reflecting real human interactions.

\paragraph{Modern Extensions}
Recent works have extended these classic models into neural settings. For example, sociologically-informed neural networks~\cite{okawa2022predicting} reinterpret opinion dynamics as a differentiable constraint in graph-based models. Similarly, the FJ model has been adapted to capture time-varying influence and diminishing competition~\cite{ballotta2024friedkin}, allowing it to explain group polarization in dynamic contexts.

\subsection{Behavior-Aware and Human-Centric Forecasting}

Traditional forecasting models primarily focus on structured components such as trend and seasonality. However, recent studies highlight that human behavioral signals — including sentiment, coordination, and crowd dynamics — also play a crucial role in shaping temporal patterns. \cite{neuman2023unveiling} reveal that collective behavior in financial markets can be quantified through ordinal pattern asymmetry and steps-to-symmetry, which measure directional imbalance and recovery time to equilibrium. Their analysis demonstrates that these behavioral distortions, often hidden in residual fluctuations, can significantly enhance the prediction of trend reversals. This implies that what has traditionally been dismissed as stochastic noise may, in fact, contain structured information reflecting human psychology.

Earlier works in behavioral economics~\cite{shiller1981stock,de1990noise} similarly emphasize that markets are influenced not only by fundamentals but also by overreaction, biased expectations, and social imitation. Building on these foundations, hybrid behavioral models~\cite{Chen2024Behavioral,Camerer2014Behavioral,Shaulska2024Theoretical,Drakopoulos2022THECOG} have sought to embed psychological and social priors into forecasting frameworks.

These findings collectively challenge the assumption that residual components are purely random. Instead, they suggest that residuals encode latent human-driven dynamics such as investor sentiment and delayed coordination. While previous methods rely on external signals like news or sentiment indices, our framework revisits this question from an endogenous viewpoint — hypothesizing that the Human Factor can be extracted directly from the intrinsic evolution of time-series residuals~\cite{Ardalani-Farsa2010Chaotic,Zhang2024Spatial,Xie2022Anomaly}.

\begin{figure*}[t]
\centering
{{\includegraphics[width=1.6\columnwidth]{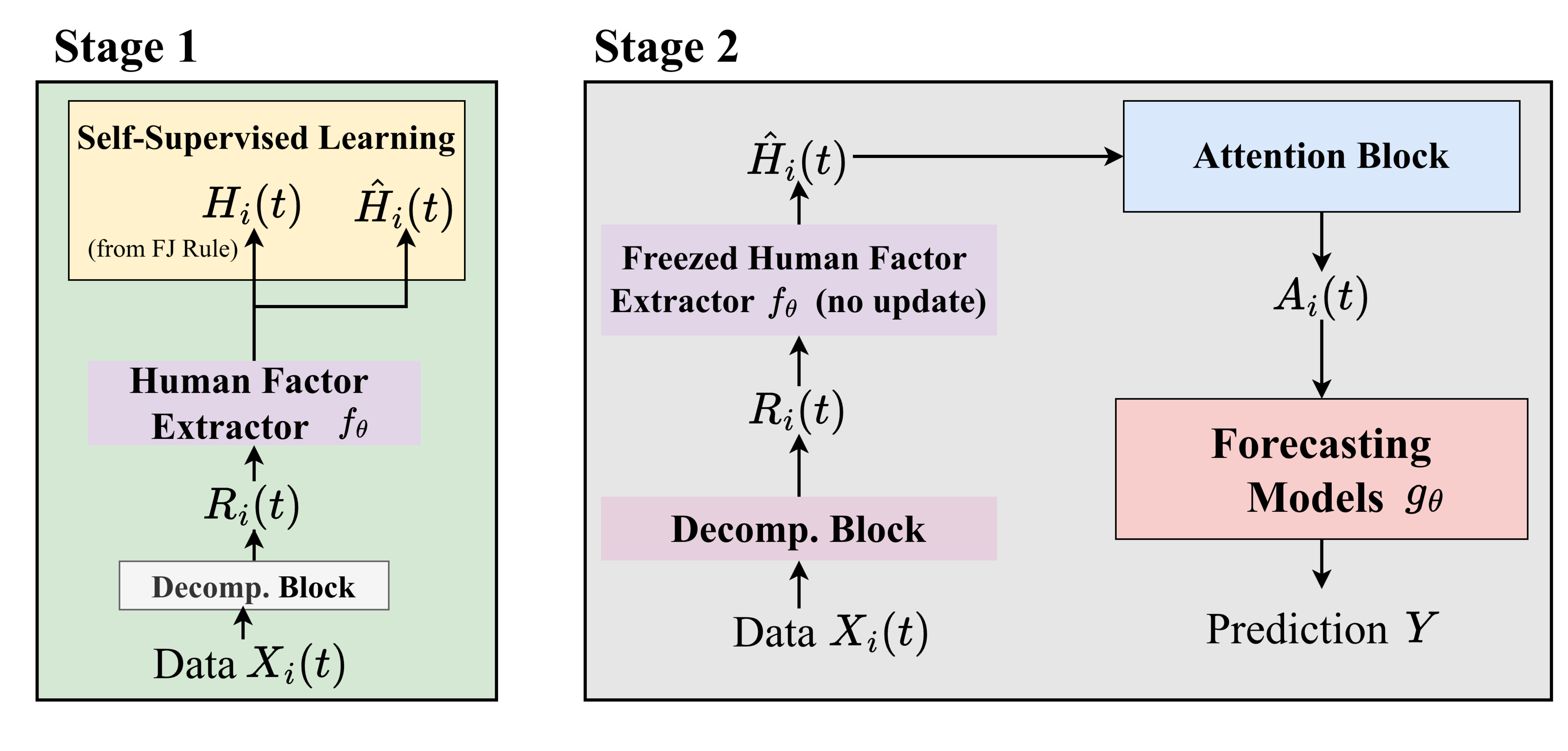}}} 
\caption{HINTS operates in two stages. In \textbf{Stage 1}, the input time series $\mathcal{X}_D^T = \{X_i(t)\}_{i,t=1}^{D,T}$ is decomposed to obtain residuals $\mathcal{R}_D^T = \{R_i(t)\}_{i,t=1}^{D,T}$, from which a Human Factor Extractor learns human-influenced signals $H_i(t)$. In \textbf{Stage 2}, the extractor is frozen and re-used to obtain $\mathcal{H}_D^T = \{H_i(t)\}_{i,t=1}^{D,T}$ from new data. This factor is passed to an attention block to produce a modulation signal $\mathcal{A}_D^T = \{A_i(t)\}_{i,t=1}^{D,T}$, which conditions the downstream forecasting model to generate the final prediction $\mathcal{Y}_D^T = \{Y_i(t)\}_{i,t=1}^{D,h}$.}
\label{fig:architecture}
\end{figure*}

\section{Proposed Method}

We propose a two-stage self-supervised framework to extract latent human behavioral signals --- termed the \textbf{Human Factor} --- from the residual component of time-series data. Our method is grounded in sociological opinion dynamics theory, specifically the Friedkin–Johnsen (FJ) model, and enables interpretable, behavior-aware forecasting \textbf{without relying on external sources}.
\begin{table}[t]
\centering
\scriptsize
\caption{Notation Summary of HINTS Framework}
\label{tbl:notation}
\setlength{\tabcolsep}{1pt}
\renewcommand{\arraystretch}{1.2}
\begin{tabular}{ll}
\toprule
\textbf{Symbol} & \textbf{Description} \\
\midrule
$\mathcal{X}^T_D = \{ X_i(t) \}_{i=1,\dots,D}^{t=1,\dots,T}$ & Input multivariate time series ($D$ variables, length $T$) \\
$\mathcal{R}^T_D = \{ R_i(t) \}$ &Residual component after STL decomposition \\
$\hat{\mathcal{H}}^T_D = \{ \hat{H}_i(t) \}$ &Extracted Human Factor from Stage 1 (latent) \\
$\mathcal{H}^T_D = \{ H_i(t) \}$& Expected Human Factor (FJ-constrained) \\
$\mathcal{A}^T_D = \{ A_i(t) \}$ &Attention map from Human Factor \\
$\tilde{\mathcal{X}}^T_D$ &Modulated input series using Human Factor \\
$\mathcal{Y}^h_D = \{ Y_i(t) \}$ &Forecasted output over horizon $h$ \\
$W = [w_{ij}]$ &Correlation-based matrix between variables \\
$\beta, \delta$ &Weights for social influence and self-memory \\
$\lambda$ &Susceptibility to new residual signals \\
$B_i(t)$ &Dynamic bias: moving average of residuals \\
$L_{FJ}$ &Stage 1 self-supervised FJ loss \\
$L_{forecast}$ &Stage 2 forecasting MSE loss \\
$\gamma$ &(Stage 2) Strength of Human Factor modulation \\
\bottomrule
\end{tabular}
\end{table}

\subsection{Overall Workflow}

As illustrated in Figure~\ref{fig:architecture}, our proposed framework HINTS processes a given time-series input $X_i(t) \in\mathbb{R}^{d}$ through the following stages:

\begin{enumerate}
    
    \item \textbf{Stage 1: Decomposition Block} We decompose the raw time-series into trend, seasonality, and residual components. The residual $R_i(t)$ serves as the primary signal of interest, hypothesized to contain latent human-driven dynamics.
    \item \textbf{Stage 1: Human Factor Extraction.} The residual $R_i(t)$ is passed to a Human Factor Extractor—a neural module constrained by the Friedkin–Johnsen opinion dynamics model. This module captures temporal patterns that reflect social influence, memory, and bias, producing a learned representation $\hat{H}_{i}(t)$, which we term the \textbf{Human Factor}. Note that the Human Factor Extractor is trained independently in Stage 1 (see Figure~\ref{fig:architecture}) and remains frozen during Stage 2.
    \item \textbf{Stage 2: Attention Network} We further process $\hat{H}_{i}(t)$ via an attention mechanism that selects and scales relevant behavioral signals, resulting in $A_i(t)$. This component acts as a modulator that highlights human-relevant dynamics.
    \item \textbf{Stage 2: Forecast Integration.} The modulated signal $A_i(t)$ is fused with the original input $X_i(t)$ and passed to any backbone forecasting model (e.g., DLinear, PatchTST, TimeMixer). This integration improves the model's ability to account for latent human factors without relying on external variables.
\end{enumerate}

This modular design enables flexible integration with existing models and allows HINTS to operate in a fully self-supervised and endogenously driven manner.

\subsection{Rationale for Using the FJ Model}\label{section:rationale}
Our goal is to extract latent dynamics ($\hat{H}$), termed the Human Factor, from the residual component. To do this, we hypothesize these dynamics are governed by the interplay of two key forces defined in our model:
\begin{enumerate}
\item Self-Dynamics (Self-Memory and Dynamic Bias terms): The tendency of the system to maintain its existing state or intrinsic bias.
\item Social Influence (Social Influence term): The change driven by impacts from other variables or external shocks.
\end{enumerate}
We adopt the Friedkin-Johnsen (FJ) opinion dynamics model because its structure is specifically designed to mathematically balance these two hypothesized forces. Unlike simpler consensus models (e.g., DeGroot), it allows for both Social Influence (Equation.~\ref{eq:social_influence}) and the retention of Self-Dynamics (Eq.~\ref{eq:self-memory} and \ref{eq:dynamic_bias}).

Therefore, HINTS leverages the FJ model not to directly measure human psychology, but because it provides the most suitable structural inductive bias to model the core properties—Social Influence, Self-Memory, and Dynamic Bias—of the latent dynamics within the residuals. We employ this FJ constraint as the self-supervised loss ($\mathcal{L}_{FJ}$) to compel the extractor ($f_{\theta}$) to learn latent factors consistent with this structure.

\subsection{Stage 1: Human Factor Extraction via Friedkin-Johnsen Constraint}

Stage 1 focuses solely on extracting a latent representation of human behavioral influence --- termed the \textbf{Human Factor} --- without using any labels or forecasting targets. Unlike traditional time-series models that learn to predict future values, this stage is purely self-supervised and optimized to satisfy a behaviorally plausible dynamic constraint.

We begin by decomposing the input time-series $\mathcal{X}_D^T = \{X_i(t)\}_{i,t=1}^{D,T}$ into trend, seasonal, and residual components using STL decomposition. Among these, the residuals, denoted $R_i(t)$, are hypothesized to contain latent, human-driven structure~\cite{neuman2023unveiling}, such as delayed reactions, herding behavior, and irrational sentiment shifts --- patterns not captured by structural decomposition.

To extract meaningful patterns from $R_i(t)$, we employ a lightweight neural network $f_\theta$ (e.g., a linear layer) to generate the Human Factor, a latent embedding defined for each variable $i$ and time step $t$ as follows:
\begin{align}
\begin{split}
    \hat{H}_i(t) := f_\theta(R_i(t)).
\end{split}
\end{align}

To guide the evolution of $\hat{H}_i(t)$, we impose a dynamic constraint inspired by the Friedkin–Johnsen (FJ) opinion dynamics model. 

The expected Human Factor update at time $t$ is defined as:

\begin{align}
\scriptsize
    H_i(t)= \underbrace{\beta \sum_{j=1}^{N} \mathbf{w}_{ij} \cdot [\lambda R_j(t-1) + (1 - \lambda) \hat{H}_j(t-1)]}_{\text{Social Influence}} \label{eq:social_influence} \\ 
   + \underbrace{\delta \cdot [\lambda R_i(t-1) + (1 - \lambda) \hat{H}_i(t-1)]}_{\text{Self-Memory}} \label{eq:self-memory}\\ 
   + \underbrace{(1 - \beta - \delta) B_i(t)}_{\text{Dynamic Bias}} \label{eq:dynamic_bias},
\end{align}

where $\mathbf{w}_{ij}$ is a correlation-based weight indicating the influence of variable $j$ on $i$, $\lambda \in [0, 1]$ is a fixed coefficient controlling susceptibility to new signals, and $B_i(t) = \frac{1}{W} \sum_{\tau = t-W}^{t-1} R_i(\tau)$ is a rolling mean or recent residuals, representing a slow-moving bias component. 
\begin{enumerate}
\item \textbf{Social Influence}: Eq.~\ref{eq:social_influence} aggregates the effects of other variables' recent signals --- both residuals and latent factors --- weighted by their relevance to $i$. It reflects how the behavior of related variables shapes the evolution of $i$'s Human Factor.
\item \textbf{Self-Memory}: Eq.~\ref{eq:self-memory} captures temporal inertia by combining $i$'s own recent residuals and past latent state. It ensures continuity and stability in the evolution of the Human Factor over time.
\item \textbf{Dynamic Bias}: Eq.~\ref{eq:dynamic_bias} introduces a local baselines from the moving average of residuals, stabilizing updates and allowing for slow, bias-driven shifts i behavior not explained by peer influence or memory alone.
\end{enumerate}

\paragraph{Self-Supervised Learning in Stage 1.} To encourage alignment with this sociologically informed update rule, we define the following self-supervised loss:
\begin{align}
    \mathcal{L}_{\text{FJ}} = \| \mathcal{H}_D^T - \hat{\mathcal{H}}_D^T \|^2\,
\end{align}
For the entire sequence across time, we denote the trajectory as $\mathcal{H}_D^T = \{ H_i(t) \}_{t,i=1}^{T,D}$, representing the temporal evolution of human influence for variable $i$. $D$ is the number of variables, and $T$ is the length of the time-series.
This formulation enables the model to learn psychologically plausible temporal dynamics without relying on ground-truth labels or external data. Stage 1 thus functions as a fully self-supervised representation learner for human-driven variations in time-series residuals. 

By minimizing this loss, the model learns to structure its internal representations in a way that is consistent with known patterns of social influence and temporal inertia.

To simplify notation in Stage 2, we denote the full sequence of extracted Human Factors as $\mathcal{H}_D^T = \{ H_i(t) \}_{i,t=1}^{D,T}$. While Stage 1 focuses on modeling fine-grained variable-wise dynamics using $H_i(t)$, Stage 2 adopts this compact representation to improve clarity, as the Human Factor is treated as a fixed, sequence-level input.

\subsection{Stage 2: Forecasting with Gated Human Modulation}

In this stage, we integrate the learned Human Factor from Stage 1 into a forecasting model to enhance its ability to capture human-driven variations. Since not all behavioral signals are equally relevant to the prediction task at each time step, we introduce a soft-attention mechanism to selectively modulate the Human Factor.

Let $\mathcal{X}_D^T \in \mathbb{R}^{D \times T}$ denote the input time-series and $\hat{\mathcal{H}}_D^T \in \mathbb{R}^{D \times T}$ the extracted Human Factor from Stage 1. We compute an attention map as:
\begin{align}
    \mathcal{A}_D^T = k_\theta(\hat{\mathcal{H}}_D^T), \label{eq:attn}
\end{align}
where $k_\theta$ is a lightweight convolutional attention network (e.g., Conv1D-Tanh-Softmax) applied over time.

The attention map is then used to modulate the input time-series as follows:
\begin{align}
    \tilde{\mathcal{X}}_D^T = \mathcal{X}_D^T + \gamma \cdot (\mathcal{X}_D^T \odot \mathcal{A}_D^T), \label{eq:modulated_input}
\end{align}
where $\gamma \in [0,1]$ is a hyperparameter that controls the strength of the modulation.

Finally, the forecasting output is obtained as:
\begin{align}
    \hat{\mathcal{Y}}_D^h = g_\theta(\tilde{\mathcal{X}}_D^T), \label{eq:forecast}
\end{align}
where $\hat{\mathcal{Y}}_D^h \in \mathbb{R}^{D \times h}$ is the forecast over the future horizon $h$, and $g_\theta$ is the forecasting model (e.g., DLinear, PatchTST, and TimeMixer).

This soft-attention mechanism enables the model to amplify or suppress specific human-driven patterns in a data-adaptive manner, improving its ability to capture behavior-induced temporal variations.

\paragraph{Model Integration.} This modulation block can be flexibly inserted into a wide range of forecasting backbones, including linear models (e.g., DLinear), transformer-based models (e.g., PatchTST), and mixing-based architectures (e.g., TimeMixer). In practice, we place the modulation block prior to the final prediction head, allowing the Human Factor to guide the model without disrupting its original structure.

\paragraph{Training.} The entire framework is trained in 2 stage, the forecasting loss $\mathcal{L}_{\text{forecast}}$:

\begin{align}
   \mathcal{L}_{\text{forecast}} = \text{MSE}(\hat{\mathcal{Y}}_D^h, \mathcal{Y}_D^h)
\end{align}
This framework enables the discovery of meaningful latent human factors embedded in residuals, offering improved forecasting accuracy and socio-psychological interpretability.

This joint training allows the extracted Human Factor to be directly optimized for downstream forecasting performance, while still preserving its interpretability via the opinion dynamics constraint.

\section{Experiments}
\label{sec:experiments}
In this section, we describe our experimental environments and results. We conduct experiments with time-series forecasting.

\subsection{Datasets}
We evaluate our model on public benchmark datasets where human factors may implicitly affect the signal and to evaluate HINTS on explicitly human-influenced domains, we employ 3 financial time-series datasets strongly affected by investor sentiment and macroeconomic behavior. Each dataset spans \textbf{January 2020–April 2025} and includes five variables: Open, High, Low, Close, and Volume.
\begin{itemize}
    \item \textbf{National Illness} \cite{cdc_fluview}: National Illness dataset, derived from the CDC's Influenza Surveillance Reports, covering the period from 2002 to 2021
    \item \textbf{Exchange Rate}~\cite{lai2018modeling}: This dataset contains daily exchange rates of eight foreign countries against the US dollar, spanning from 1990 to 2016. 
    \item \textbf{Traffic}~\cite{caltrans_pems} : Traffic dataset represents the road occupancy rates, capturing hourly data recorded by sensors on the San Francisco freeways between 2015 and 2016. 
    \item \textbf{Electricity} \cite{godahewa2020electricity}: Electricity dataset tracks the hourly electricity consumption of 321 clients from 2012 to 2014.
    \item \textbf{PeMS} \cite{caltrans_pems}: a spatio-temporal traffic speed dataset collected by Caltrans District 3 in California during September 2018. It contains traffic speed readings at 5-minute intervals from 358 sensors. 
    \item \textbf{Tech Stocks:} Daily stock data from major technology firms, reflecting behavioral trends within the tech sector.
    \item \textbf{S\&P 100 / S\&P 500:} Broad U.S. market indices representing collective investor behavior across industries and market scales.
\end{itemize}

\subsection{Baselines}
We compare our method against the following baselines:

We consider three strong baselines that operate solely on raw time-series inputs without incorporating external sources or latent factors:
\begin{itemize}
\item DLinear~\cite{zeng2023transformers} is a decomposition-based linear forecasting model that separately learns trend and seasonal components. 
\item PatchTST~\cite{nie2022time} applies a Transformer architecture over time-series patches, enabling efficient long-range dependency modeling while capturing local structures.
\item TimeMixer~\cite{wang2024timemixer} leverages a mixing architecture inspired by MLP-Mixer to model temporal dynamics through channel-wise and temporal-wise mixing. 
\item{LLM-based External Modeling.} We compare to \textbf{From News to Forecast} \cite{wang2024news}, a recent method that integrates filtered news content into LLM-based forecasters in Section~\ref{ablation}. 
\end{itemize}

\subsection{Hyperparameters}
For reproducibility, we specify the hyperparameter search spaces used in our experiments. We tune two learning rates independently: $\eta_{\theta_f}$ for the Human Factor extractor in Stage 1, and $\eta_{\theta_g}$ for the forecasting backbone in Stage 2. Both are selected from $\{10^{-2}, 10^{-3}, 5 \times 10^{-4}, 10^{-4}\}$ based on validation performance. In Stage 2, the weighting parameter $\gamma$ for incorporating the Human Factor is tuned from $\{0.1, 0.3, 0.5, 0.9, 1.0\}$. All other architectural and optimization parameters follow the default settings of each backbone model (DLinear, PatchTST, TimeMixer).

\subsection{Main Results}
We propose HINTS, a forecasting framework that extracts Human Factors from time-series data and leverages them to guide predictions via attention-based modulation.
Across benchmark datasets (Tables~\ref{tbl:parta1}), HINTS consistently outperforms all backbones and horizons, achieving up to 28.9\% improvement on PEMS traffic data by modeling collective mobility patterns, and 12.7\% on the Exchange dataset by capturing market-level sentiment and anticipatory behavior. Even in domains with weaker behavioral effects, such as Illness, it yields stable gains (up to 32.6\%), showing strong generalization beyond explicitly social contexts.

On real-world financial datasets (Table~\ref{tbl:parta1})—including Tech Stocks, S\&P 100/500, and Retail Stocks—HINTS enhances forecasting accuracy across all models and horizons, with improvements up to 15.2\%. Gains are particularly evident at longer horizons ($h=48,60$), where compounding errors typically occur. These results demonstrate that even advanced forecasting models benefit from Human Factor–driven refinement, which captures latent behavioral dynamics such as sentiment, coordination, and herding that are not directly observable in raw price data.

\begin{table}[h]
\caption{Comparison between News to Forecast and HINTS}\label{tbl:comparison_news_to_forecast}
\renewcommand{\arraystretch}{1.2}
\setlength{\tabcolsep}{2.9pt}
\resizebox{\linewidth}{!}{%
\begin{tabular}{c cc cc cc} \toprule
Datasets  & \multicolumn{2}{c}{Traffic} & \multicolumn{2}{c}{Electricity} & \multicolumn{2}{c}{Exchange} \\
\cmidrule(lr){2-3} \cmidrule(lr){4-5} \cmidrule(lr){6-7}
Methods
& $\text{MSE}_{\times 10^{3}}$ & $\text{MAE}_{\times 10^{2}}$
& $\text{MSE}_{\times 10^{-3}}$ & $\text{MAE}$
& $\text{MSE}_{\times 10^{4}}$ & $\text{MAE}_{\times 10^{3}}$ \\
\midrule
DLinear                                  & 1.67 & 1.70   & 161.6 & 255.7  & 0.91 & 6.96 \\
PatchTST                                 & 1.54 & 1.84   & 133.5 & 234.5  & 0.77 & 6.73 \\
News to Forecast~\cite{wang2024news}      & 1.78 & 1.43   & \color{blue}{78.62} & 180.9  & \color{blue}{0.42} & \color{blue}{4.83} \\
\midrule
HINTS (DLinear)                          & \textbf{1.62} & \textbf{1.64}   & \textbf{92.13} & \textbf{176.9}  & \textbf{0.58} & \textbf{5.12} \\
HINTS (PatchTST)                         & \color{blue}{\textbf{1.38}} & \color{blue}{\textbf{0.98}}   & \textbf{80.18} & \textcolor{blue}{\textbf{167.7}}  & \textbf{0.44} & \textbf{4.91} \\
\bottomrule
\end{tabular}%
}
\end{table}

\begin{figure}[t]
\begin{center}
\subfigure[AMZN]{{\includegraphics[width=0.92\columnwidth]{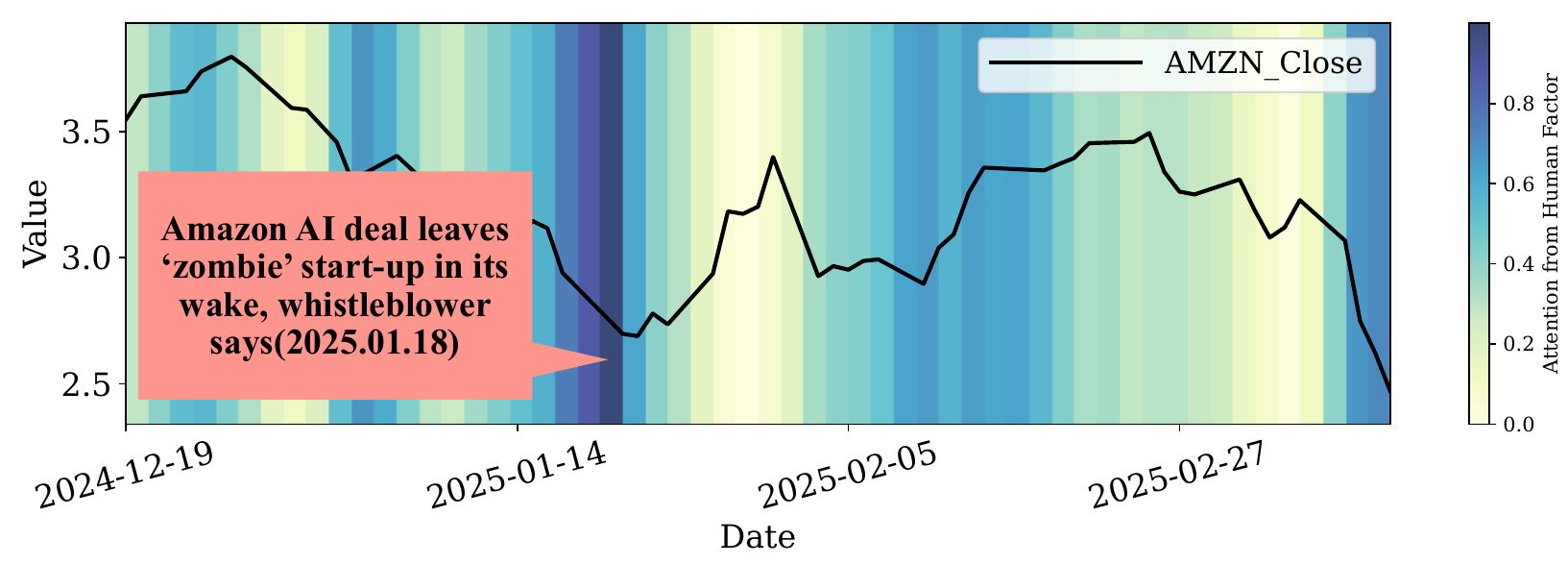}}}\\
\subfigure[NVDA]{{\includegraphics[width=0.92\columnwidth]{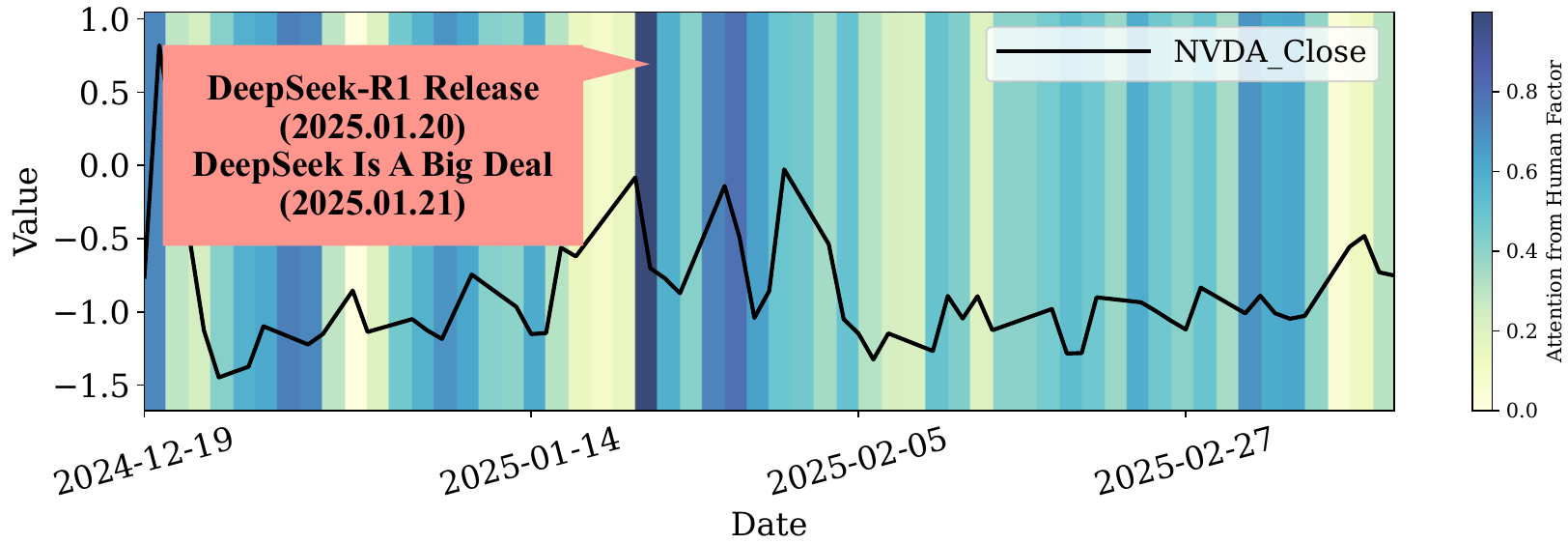}}}\\
\subfigure[GOOGL]{{\includegraphics[width=0.92\columnwidth]{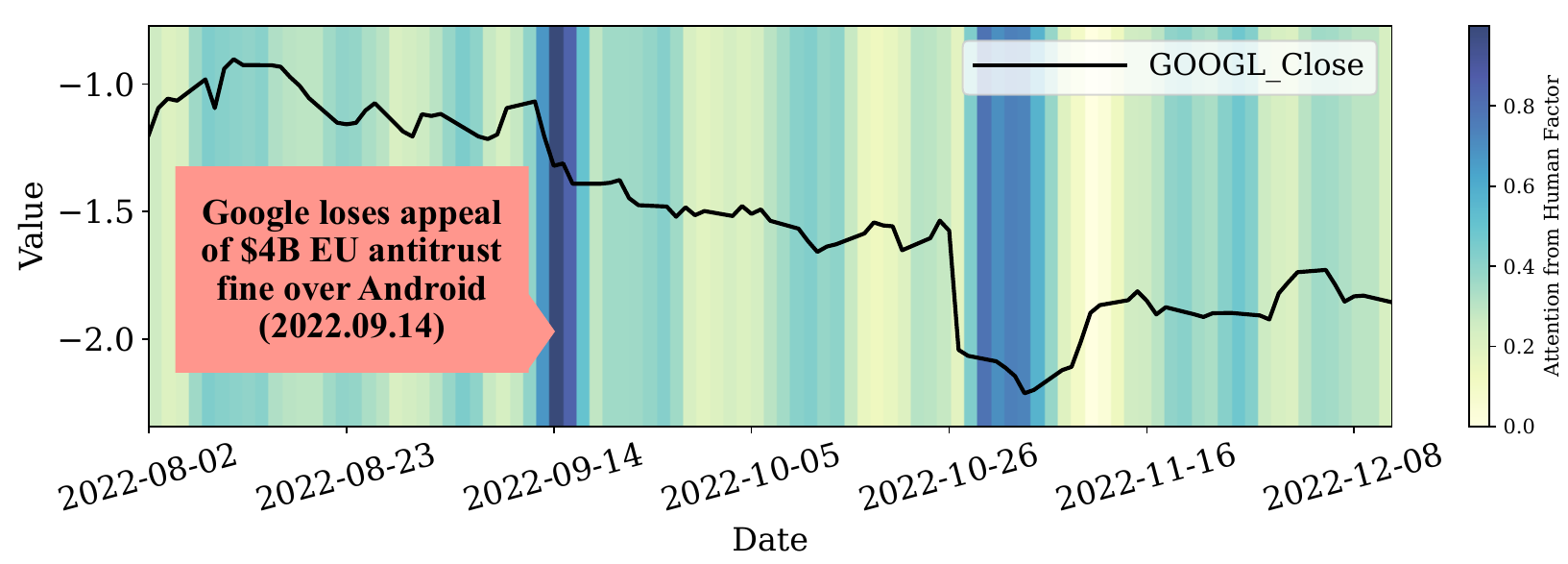}}} 
\caption{ Visualization of the extracted Human Factor $\hat{\mathcal{H}_D^T}$ based attention $\mathcal{A}_D^T$ for major tech stocks.}\label{fig:vis1}
\end{center}
\end{figure} 
\vspace{-2em}

\begin{table*}[t]
\scriptsize
\caption{Experimental results on 9 benchmark datasets. Results where the HINTS method improved the performance of a base model are highlighted in \textbf{bold}, and the best performance w.r.t. a pair of (dataset, horizon $h$) among all is marked in \textcolor{blue}{\textbf{blue}}. $h$ refers to the prediction horizon.}\label{tbl:parta1}
\setlength{\tabcolsep}{2.7pt}
\renewcommand{\arraystretch}{0.95}
\begin{tabular}{ccccccccccccccccccccccc} \toprule
\multicolumn{2}{c}{Datasets}  & \multicolumn{2}{c}{Exchange}  && \multicolumn{2}{c}{ILL} & \multicolumn{2}{c}{Tech Stock 10}  & \multicolumn{2}{c}{S\&P 100} & \multicolumn{2}{c}{S\&P 500} && \multicolumn{2}{c}{PeMS03} & \multicolumn{2}{c}{PeMS04} & \multicolumn{2}{c}{PeMS07}  & \multicolumn{2}{c}{PeMS08} \\  \cmidrule(lr){3-4} \cmidrule(lr){6-7} \cmidrule(lr){8-9} \cmidrule(lr){10-11}  \cmidrule(lr){12-13}  \cmidrule(lr){15-16} \cmidrule(lr){17-18} \cmidrule(lr){19-20} \cmidrule(lr){21-22} 
 & $h$   & MSE & MAE & $h$ & MSE & MAE & MSE & MAE & MSE & MAE & MSE & MAE & $h$ & MSE & MAE  & MSE & MAE  & MSE & MAE & MSE & MAE \\\midrule
\multirow{4}{*}{\begin{sideways}DLinear\end{sideways}} 
\multirow{4}{*}{\begin{sideways}(2023)\end{sideways}} 
& 96   & 0.081 & 0.203 &24 & 2.215 & 1.081           & 0.732 & 0.547  & 0.781 & 0.594 & 0.781 & 0.503 & 12  & 0.122 & 0.243 & 0.148 & 0.272 & 0.115 & 0.242 & 0.154 & 0.276     \\
& 192  & 0.157 & 0.293 &36 & 1.963 & 0.963          & 0.839 & 0.591  & 1.004 & 0.682 & 0.956 & 0.570 & 24  & 0.201 & 0.317 & 0.224 & 0.340 & 0.210 & 0.329 & 0.248 & 0.353     \\
& 336  & 0.305 & 0.414 &48 & \textbf{2.130} & 1.024  & 1.482 & 0.827  & 1.246 & 0.760 & 1.436 & 0.735  & 48  & 0.333 & 0.425 & 0.355 & 0.437 & 0.398 & 0.458 & 0.440 & 0.470 \\
& 720  & 0.643 & 0.601 &60 & 2.368 & 1.096         & 1.414 & 0.797  & 1.334 & 0.771 & 1.781 & 0.825 & 96  & 0.457 & 0.515 & 0.452 & 0.504 & 0.594 & 0.553 & 0.674 & 0.565     \\\midrule
\multirow{4}{*}{\begin{sideways}HINTS\end{sideways}} 
\multirow{4}{*}{\begin{sideways} DLinear \end{sideways}} 
& 96   & \color{blue}{\textbf{0.077}} & \color{blue}{\textbf{0.198}}  &24 & \textbf{2.205} & \textbf{1.035} & \textbf{0.653} & \textbf{0.507} & \textbf{0.566} & \textbf{0.481} & \textbf{0.736} & \textbf{0.483}  & 12  & \textbf{0.104} & \textbf{0.219} & \textbf{0.114} & \textbf{0.226} & \textbf{0.099} & \textbf{0.214} & \textbf{0.111} & \textbf{0.221} \\
& 192  & \color{blue}{\textbf{0.150}} & \color{blue}{\textbf{0.284}}  &36 & \textbf{1.951}  & \textbf{0.957}& \textbf{0.819} & \textbf{0.582} & \textbf{0.757} & \textbf{0.569} & \textbf{0.871} & \textbf{0.540}  & 24  & \textbf{0.181} & \textbf{0.295} & \textbf{0.188} & \textbf{0.297} & \textbf{0.187} & \textbf{0.300} & \textbf{0.194} & \textbf{0.299} \\
& 336  & \color{blue}{\textbf{0.258}} & \color{blue}{\textbf{0.380}}  &48 & 2.132 & \textbf{1.021}          & \textbf{1.193} & \textbf{0.723} & \textbf{1.072} & \textbf{0.686} & \textbf{1.096} & \textbf{0.623}  & 48  & \textbf{0.317} & \textbf{0.409} & \textbf{0.320}      & \textbf{0.403}  & \textbf{0.373} & \textbf{0.434} & \textbf{0.384} & \textbf{0.430} \\
& 720  & \color{blue}{\textbf{0.550}} & \color{blue}{\textbf{0.579}}  &60 & \textbf{2.180} & \textbf{1.042} & \textbf{1.308} & \textbf{0.751} & \textbf{1.162} & \textbf{0.721} & \textbf{1.348} & \textbf{0.700} & 96  & \textbf{0.451} & \textbf{0.507} & \textbf{0.432} & \textbf{0.485} & \textbf{0.582} & \textbf{0.539} & \textbf{0.640} & \textbf{0.541} \\\bottomrule
Imp.(Avg.)&  & 12.7\% & 4.63\% & & 2.40\% & 2.62\%  & 12.7\% & 4.63\% & 1.80\% & 1.61\% & 0.81\% & 1.44\%  &  & 5.39\% & 4.66\% & 10.6\% & 9.14\% & 5.77\% & 6.01\% &  12.3\% & 10.4\% \\\bottomrule
\multirow{4}{*}{\begin{sideways}PatchTST\end{sideways}} 
\multirow{4}{*}{\begin{sideways}(2023)\end{sideways}} 
& 96   & 0.093 & \textbf{0.218}&24 & 1.319 & 0.754 & 0.554 & 0.460  & 0.424 & 0.401 & 0.577 & 0.406  & 12  & 0.099 & 0.216 & 0.105 & 0.224 & 0.095 & 0.207 & 0.168 & 0.232 \\
& 192  & 0.208 & 0.332  &36 & 1.007 & 0.870  & 0.752 & 0.533  & 0.510 & 0.457 & 0.678 & 0.475  & 24  & 0.142 & 0.259 & 0.153 & 0.275 & 0.150 & 0.262 & 0.224 & 0.281 \\
& 336  & 0.359 & 0.440  &48 & 1.553 & 0.815  & 0.786 & 0.564  & 0.593 & 0.494 & 0.766 & 0.502  & 48  & 0.211 & 0.319 & 0.229 & 0.339 & 0.253 & 0.340 & 0.321 & 0.354 \\
& 720  & 1.194 & 0.815  &60 & \color{blue}{\textbf{1.016}} & 0.788 & 0.953 & 0.624  & 0.648 & 0.525 & 0.866 & 0.553  & 96  & 0.269 & 0.370 & 0.291 & 0.389 & 0.346 & 0.404 & 0.408 & 0.417 \\\midrule
\multirow{4}{*}{\begin{sideways}HINTS\end{sideways}} \multirow{4}{*}{\begin{sideways}PatchTST\end{sideways}} 
& 96   & \textbf{0.090} & \textbf{0.218}  &24 & \textbf{1.207} & \textbf{0.732} & \textbf{0.500} & \textbf{0.435}  & \textbf{0.420} & \textbf{0.392} & \textbf{0.570} &  \textbf{0.396}  & 12  & \color{blue}{\textbf{0.076}} & \color{blue}{\textbf{0.182}} & \textbf{0.099} & \textbf{0.206} & \textbf{0.092} & \textbf{0.206} & \textbf{0.111} & \textbf{0.224} \\
& 192  & \textbf{0.179} & \textbf{0.313}  &36 & \color{blue}{\textbf{0.880}} & \textbf{0.642} & \textbf{0.613} & \textbf{0.485}  & \textbf{0.499} & \textbf{0.447} & \textbf{0.667} & \textbf{0.452}  & 24  &\textbf{0.127} & \textbf{0.238} & \textbf{0.149} & \textbf{0.255} & \textbf{0.118} & \textbf{0.223} & \textbf{0.141} & \textbf{0.245} \\
& 336  & \textbf{0.290} & \textbf{0.381}  &48 & \textbf{1.423} & \textbf{0.782}               & \textbf{0.696} & \color{blue}{\textbf{0.530}}  & \textbf{0.567} & \textbf{0.489} & \textbf{0.758} & \textbf{0.498}  & 48  & \textbf{0.210} & \textbf{0.307} & \textbf{0.217} & \textbf{0.314} & \textbf{0.221} & \textbf{0.308} & \textbf{0.249} & \textbf{0.325} \\
& 720  & \textbf{1.179} & \textbf{0.745}  &60 & 1.031 & \textbf{0.773}                        & \textbf{0.774} & \textbf{0.569}  & \color{blue}{\textbf{0.632}} & \color{blue}{\textbf{0.523}} & \textbf{0.854} & \color{blue}{\textbf{0.538}}  & 96  & \color{blue}{\textbf{0.258}} & \textbf{0.361} & \color{blue}{\textbf{0.279}} & \color{blue}{\textbf{0.358}} & \textbf{0.342} & \textbf{0.401} & \textbf{0.388} & \textbf{0.409} \\\bottomrule
Imp.(Avg.)&  & 6.26\% & 8.20\% & & 7.23\% & 9.23\% & 15.2\% & 7.43\% & 2.62\% & 1.39\% & 1.32\% & 2.69\%  &  & 6.94\% & 6.53\% & 4.37\% & 7.66\% & 8.41\% & 6.18\% &  20.7\% & 6.31\%  \\\bottomrule
\multirow{4}{*}{\begin{sideways}TimeMixer\end{sideways}} 
\multirow{4}{*}{\begin{sideways}(2024)\end{sideways}} 
& 96  & 0.093 & 0.212  & 24 & 1.469 & 0.798 & 0.535 & 0.447  & 0.391 & 0.387 & 0.553 & 0.389  & 12  & 0.099 & 0.205 & 0.103 & 0.213 & 0.096 & 0.187 & 0.099 & 0.211 \\
& 192 & 0.174 & 0.297  & 36 & 1.890 & 0.867 & 0.616 & 0.492  & 0.483 & 0.477 & 0.612 & 0.432  & 24  & 0.133 & 0.246 & 0.150 & 0.270 & 0.148 & 0.258 & 0.148 & 0.268  \\
& 336 & 0.349 & 0.426  & 48 & 1.885 & 0.924 & 0.762 & 0.548  & 0.571 & 0.503 & 0.683 & 0.502  & 48  & 0.208 & 0.302 & 0.226 & 0.332 & 0.258 & 0.351 & 0.237 & 0.387  \\
& 720 & 1.065 & 0.770  & 60 & 1.955 & 0.980 & 0.866 & 0.607  & 0.689 & 0.558 & 0.773 & 0.631  & 96  & 0.352 & 0.391 & 0.288 & 0.381 & 0.343 & 0.388 & 0.385 & 0.402  \\\midrule
\multirow{4}{*}{\begin{sideways}HINTS\end{sideways}} \multirow{4}{*}{\begin{sideways}TimeMixer\end{sideways}} 
& 96   & \textbf{0.090} & \textbf{0.211} &24 & \color{blue}{\textbf{1.135}} & \color{blue}{\textbf{0.558}}  & \color{blue}{\textbf{0.481}} & \color{blue}{\textbf{0.422}}  & \color{blue}{\textbf{0.380}} & \color{blue}{\textbf{0.373}} & \color{blue}{\textbf{0.502}} & \color{blue}{\textbf{0.329}}  & 12  & \textbf{0.080} & \textbf{0.192} & \color{blue}{\textbf{0.088}} & \color{blue}{\textbf{0.200}} & \color{blue}{\textbf{0.075}} & \color{blue}{\textbf{0.182}} & \color{blue}{\textbf{0.081}} & \color{blue}{\textbf{0.194}} \\
& 192  & \textbf{0.170} & \textbf{0.294} &36 & \textbf{1.149} & \color{blue}{\textbf{0.604}}             & \color{blue}{\textbf{0.586}} & \color{blue}{\textbf{0.480}}  & \color{blue}{\textbf{0.462}} & \color{blue}{\textbf{0.432}} & \color{blue}{\textbf{0.582}} & \color{blue}{\textbf{0.408}}  & 24  & \color{blue}{\textbf{0.111}} & \color{blue}{\textbf{0.223}} & \color{blue}{\textbf{0.116}} & \color{blue}{\textbf{0.232}} & \color{blue}{\textbf{0.098}} & \color{blue}{\textbf{0.213}} & \color{blue}{\textbf{0.118}} & \color{blue}{\textbf{0.232}} \\
& 336  & \textbf{0.339} & \textbf{0.418} &48 & \color{blue}{\textbf{1.235}} & \color{blue}{\textbf{0.635}}  & \color{blue}{\textbf{0.688}} & \textbf{0.531}  & \color{blue}{\textbf{0.551}} & \color{blue}{\textbf{0.484}} & \color{blue}{\textbf{0.627}} & \color{blue}{\textbf{0.493}}  & 48  & \color{blue}{\textbf{0.169}} & \color{blue}{\textbf{0.287}} & \color{blue}{\textbf{0.183}} & \color{blue}{\textbf{0.293}} & \color{blue}{\textbf{0.192}} & \color{blue}{\textbf{0.295}} & \color{blue}{\textbf{0.175}} & \color{blue}{\textbf{0.289}} \\
& 720  & \textbf{0.916} & \textbf{0.717} &60 & \textbf{1.332} & \color{blue}{\textbf{0.682}}           & \color{blue}{\textbf{0.720}} & \color{blue}{\textbf{0.563}}  & \textbf{0.658} & \textbf{0.538} & \color{blue}{\textbf{0.703}} &\textbf{0.608}  & 96  & \textbf{0.264} & \color{blue}{\textbf{0.352}} & \textbf{0.281} & \textbf{0.368} & \color{blue}{\textbf{0.290}} & \color{blue}{\textbf{0.365}} & \color{blue}{\textbf{0.243}} &  \color{blue}{\textbf{0.336}} \\\bottomrule
Imp.(Avg.)&  & 9.87\% & 3.81\% & & 32.6\% & 30.5\% & 10.9\% & 4.68\% & 3.89\% & 5.09\% & 7.89\% & 5.93\%  &  & 21.2\% & 7.87\% & 12.9\% & 8.61\% & 22.5\% & 10.9\% &  28.9\% & 17.1\%  \\\bottomrule
\end{tabular}
\end{table*}

\subsection{Case Study and Sensitivity Analysis}\label{ablation}

\paragraph{Human Factor Interpretability}

Figure ~\ref{fig:vis1} visualizes the human factors attention map for major technology stocks (AMZN, GOOGL, NVDA). The lines represent closing prices, and the background represents actual news events annotated with Human Factors extracted from HINTS. Despite being trained solely on time-series data, the extracted Human Factors closely align with significant market trends (often occurring one to two days in advance). This demonstrates that the extracted Human Factors capture potential sentiment and behavioral shifts without external input. These results demonstrate the potential of HINTS to provide interpretable, behaviorally informed financial forecasts, even without explicit news or sentiment data.

\paragraph{Ablation Studies on HINTS Components}
To validate the individual contributions of our key components, as hypothesized in our Section~\ref{section:rationale}, we conducted an ablation study. The results are presented in Table~\ref{tbl:ablation} (PeMS 08, $h=720$, TimeMixer). We observe that the HINTS (Baseline, w/o $\mathcal{L}_{FJ}$) variant, which removes our entire constraint, shows the highest error, confirming the necessity of our self-supervised approach. Furthermore, removing either the Social Influence component (defined in Eq.~\ref{eq:social_influence}) or the combined Self-Memory and Dynamic Bias components (from Eq. ~\ref{eq:self-memory} and ~\ref{eq:dynamic_bias}) individually also resulted in a significant degradation in performance. This experimentally validates our hypothesis that both Social Influence and the combined Self-Memory and Dynamic Bias components are essential and contribute individually to the model's forecasting accuracy.

\begin{table}[h]
\scriptsize
\centering
\caption{Ablation study on the components of the HINTS framework. We compare the full model against variants with key components removed.}
\label{tab:ablation}
\begin{tabular}{@{}ccc@{}}
\toprule
\textbf{Model Variant} & \textbf{MSE (↓)} & \textbf{MAE (↓)}  \\
\midrule
HINTS (Full Model) & \textbf{0.243} & \textbf{0.336}  \\
HINTS w/o Social Influence & 0.325 & 0.375  \\
HINTS w/o (Self-Memory, Bias) & 0.342 & 0.401  \\
HINTS (Baseline, w/o $\mathcal{L}_{FJ}$) & 0.385 & 0.402  \\
\bottomrule
\end{tabular} \label{tbl:ablation}
\end{table}

\paragraph{Comparison between News to Forecast and HINTS}
From News to Forecast~\cite{wang2024news} (hereafter \textit{News to Forecast}) leverages rich exogenous sources such as external news and weather data, whereas the HINTS framework operates strictly in an endogenous setting without any external inputs. Table~\ref{tbl:comparison_news_to_forecast} reproduces the \textit{News to Forecast} setup for a fair comparison.
Despite this disadvantage, HINTS consistently improves over the backbones (e.g., DLinear, PatchTST). In the Traffic dataset, \textit{News to Forecast} offers little gain over the original models, suggesting limited benefit from exogenous variables, yet HINTS still yields notable improvements --- capturing latent behavioral signals within the series. For datasets like Electricity and Exchange, where exogenous variables significantly boost performance, HINTS, without any external data, delivers effects comparable to those variables. Remarkably, across several settings, HINTS achieves accuracy levels that rival \textit{News to Forecast}, demonstrating that its self-supervised human factor extraction can recover much of the predictive advantage typically gained from exogenous sources, highlighting its value when external information is costly, inaccessible, or noisy.

\begin{figure}[t]
\begin{center}
\subfigure[Exchange($h$=336)]{{\includegraphics[width=0.47\columnwidth]{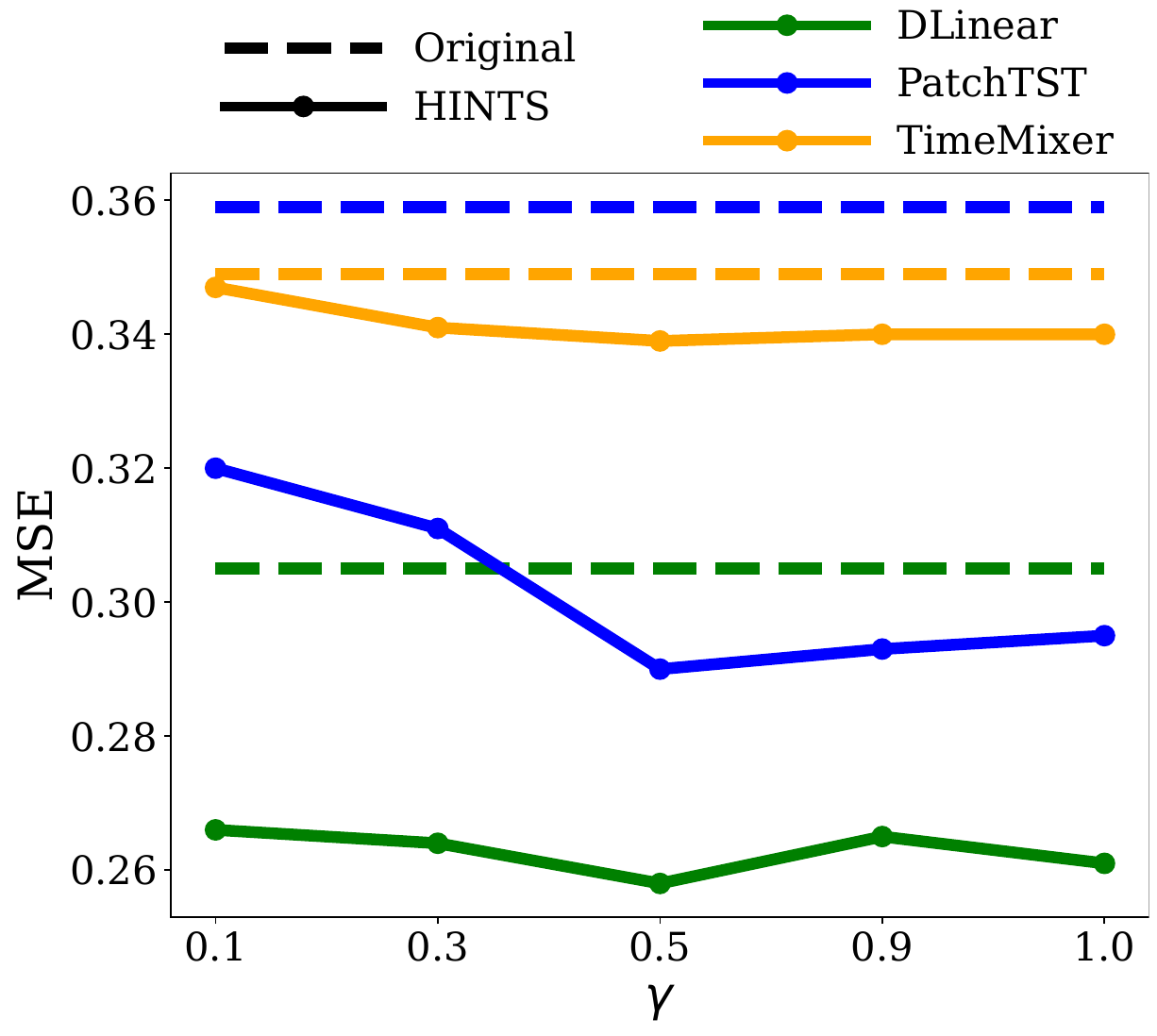}}} \hfill
\subfigure[PEMS04 ($h$=24)]{{\includegraphics[width=0.47\columnwidth]{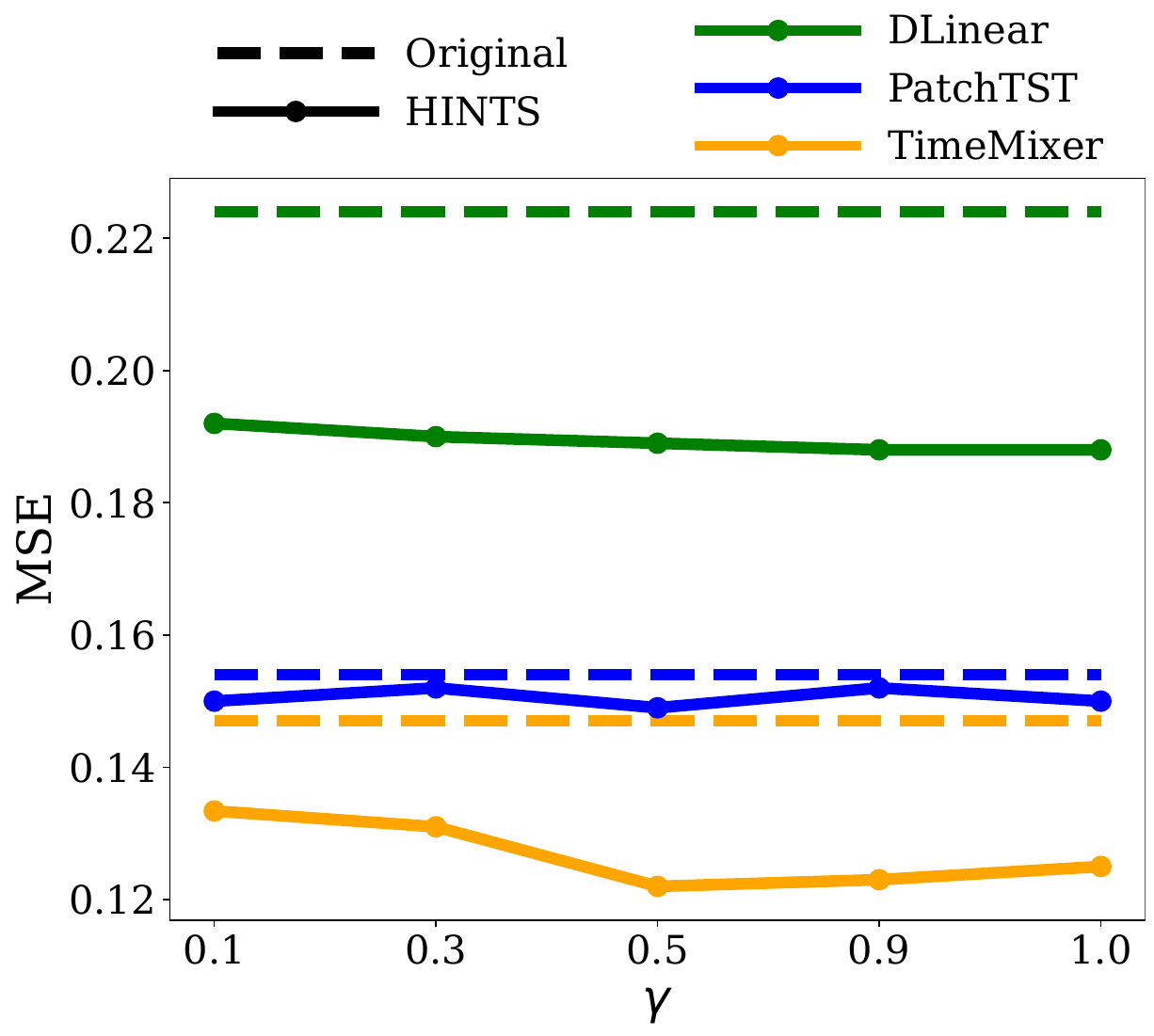}}} 
\caption{Sensitivity analysis with respect to the Human Factor weighting parameter $\gamma$ in Stage 2 of HINTS.}\label{fig:sensitivity1}
\end{center}
\end{figure}

\paragraph{Sensitivity to the Human Factor Weight $\gamma$}
In our framework, the parameter $\gamma$ determines how strongly the extracted Human Factor $\hat{\mathcal{H}}_D^T$ influences the forecasting input. Specifically, $\gamma$ controls the strength of the attention matrix $\mathcal{A}_D^T$, derived from $\hat{\mathcal{H}}_D^T$, in adjusting the original input $\mathcal{X}_D^T$. As $\gamma$ increases, the model places greater emphasis on the human-driven behavioral signals captured by the attention. As shown in Figure~\ref{fig:sensitivity1}, we observe that larger $\gamma$ values generally lead to improved forecasting performance across both Exchange and PEMS04 datasets. This confirms that the Human Factor --- driven attention provides meaningful cues, and amplifying its contribution improves the model’s predictive accuracy.

\section{Conclusion}
In this work, we introduced HINTS, a self-supervised framework designed to extract latent human-driven dynamics --- termed the Human Factor --- directly from time-series residuals without relying on any external sources such as news articles, social media, or sentiment indices. Unlike previous approaches that depend on external data pipelines, HINTS identifies meaningful behavioral patterns endogenously by modeling opinion dynamics from within the data itself. Through extensive experiments across diverse domains --- including traffic, energy, and finance --- we demonstrated that residual signals, traditionally treated as noise, often encode structured human influence. By leveraging this insight, HINTS improves both forecasting accuracy and interpretability. These findings suggest that modeling internal behavioral signals offers a robust and scalable alternative to externally driven methods, particularly in scenarios where external sources are limited, costly, or unavailable.

\section{Acknowledgments}

This work was supported by Institute of Information \& communications Technology Planning \& Evaluation (IITP) grant funded by the Korea government (MSIT) (No.2022-0-00857, Development of Financial and Economic Digital Twin Platform based on AI and Data)


\end{document}